# REVISITING FACIAL KEY POINT DETECTION: AN EFFICIENT APPROACH USING DEEP NEURAL NETWORKS

* Prathima Dileep[1], Bharath Kumar Bolla[2], Sabeesh Ethiraj[3]

[1]Upgrad Education Pvt. Ltd., Mumbai, India
[1]prathi8dec@gmail.com

[2]Salesforce, Hyderabad, India
[2]bolla111@gmail.com

[3]Liverpool John Moore University, London
[3]sabeesh90@yahoo.co.uk

**Abstract** Facial landmark detection is a widely researched field of deep learning as this has a wide range of applications in many fields. These key points are distinguishing characteristic points on the face, such as the eyes center, the eye's inner and outer corners, the mouth center, and the nose tip from which human emotions and intent can be explained. The focus of our work has been evaluating transfer learning models such as MobileNetV2 and NasNetMobile, including custom CNN architectures. The objective of the research has been to develop efficient deep learning models in terms of model size, parameters, and inference time and to study the effect of augmentation imputation and fine-tuning on these models. It was found that while augmentation techniques produced lower RMSE scores than imputation techniques, they did not affect the inference time. MobileNetV2 architecture produced the lowest RMSE and inference time. Moreover, our results indicate that manually optimized CNN architectures performed similarly to Auto Keras tuned architecture. However, manually optimized architectures yielded better inference time and training curves.

**Keywords:** Inference Time, Efficient Transfer Learning, Deep Learning, MobileNetV2, NasNetMobile, Custom CNN, Keras-autotuner, Augmentation

## 1  Introduction

The face is critical in visual communication. Numerous nonverbal messages, such as human identity, intent, and emotion, can be automatically extracted from the face. Localizations of facial key points are required in computer vision to extract nonverbal cues of facial information automatically. The term "facial appearance" refers to the distinct patterns of pixel intensity around or across facial landmarks or key points. These key points represent those critical features on a human face, such as the eyes, nose, eyebrows, lips, and nose from which information about a person's emotion or intent can be identified. Once correctly identified, they can be used to train deep learning algorithms to perform various classification tasks. Their applications include computer interaction, entertainment, drowsiness detection, biometrics, emotion detection, security surveillance, and a range of medical applications. However, the practical applications of these models depend on the speed of inference of these models and their deployability on Edge and mobile devices that have lower computational powers. This research aims to evaluate various transfer learning and custom models in terms of inference time, model size to test their deployability on Edge / mobile devices.

In this work, we used the Facial Key Point Detection dataset from Kaggle. The dataset consists of the training variables and 15 target variables, the facial key points representing various facial features. Deep learning models using custom and transfer learning architectures such as Resnet50, MobileNetV2, NasnetMobile have been built using baseline and by combining various augmentation techniques to identify the ideal model. Additionally, the architectures have been evaluated in terms of parameter count, disc requirements, and inference timings to determine their suitability for deployment on computationally less intensive devices. We have compared our results with other state-of-the-art architectures and found that our models have higher efficiency, hence achieving the objective of this research.

## 2 Literature Review

Facial landmark detection algorithms can be classified into three broad categories [1] based on how they model the facial appearance and shape: holistic, Constrained Local Model (CLM), and regression based. Holistic methods mainly include Active Appearance Models (AAM) [2] and fitting algorithms. AAM works on the principle of learning from the whole face patch and involves the concept of PCA, wherein learning takes place by calculating the difference "I" between the greyscale image and an instance of the model. The error is reduced by learning the parameters like any conventional machine learning algorithm. CLM methods are slightly better than the holistic approaches as they learn from both the globalized face pattern and the local appearance from the nearby facial key points. They can be probabilistic or deterministic. They consist of two steps [3], the initial step where the landmarks are located independent of the other landmarks. In this second step, while updating the parameters, the location of all the landmarks is updated simultaneously. In regression-based approaches, there is no initial localization of the landmark; instead, the images are mapped directly to the co-coordinates of these landmarks, and the learning is done directly. These methods may be direct or cascaded. However, with the advent of deep learning algorithms, convolutional neural networks have replaced conventional regression methods with state-of-the-art results. These methods are faster and more efficient. Convolutional neural networks using LeNet have been used in many state-of-the-art works. The principles of LeNet have been used to build many custom architectures, which have shown reduced training time [4] and reduced RMSE scores.

The performance of a machine learning model also depends mainly on the type of algorithm being used. Some of the popular datasets [1] on which deep learning algorithms have been used with promising results are BU-4DFE with 68 landmark points (RMSE – 5.15), AFLW with 53 landmark points (RMSE-4.26), AFW with five landmark points (RMSE – 8.2), LFPW with 68 landmark points (RMSE – 5.44), Ibug 300-W with 67 landmark points (RMSE – 5.54). Most of the deep learning algorithms have utilized methods such as Task constrained deep convolutional network (TCDCN) [5], Hyperface [6], 3-Dimensional, Dense Face Alignment (3DDFA) [7], Coarse to Fine Auto Encoder Techniques (CFAN) [8] in achieving relatively higher accuracies.

Inception architecture [9] has been used on a similar Kaggle dataset achieving an RMSE score of 2.91. Resnet has also been used in work done by [10], achieving an RMSE of 2.23. Similar work done using the LeNet architecture [4] achieved an RMSE score of 1.77. As more evidence was produced favouring custom architectures, the focus was directed to build Custom CNN networks for facial key point detection. A comparative study was done [11] using both custom and transfer learning architectures. Custom architectures were able to achieve lower RMSE scores (1.97). A similar custom model consisting of 14 layers [12] produced an RMSE score of 1.75. As evident above, attaining higher accuracy by making deep learning algorithms more efficient and precise has been the target of various studies.

Tuning of deep learning models is also critical in achieving high accuracies. The Keras tuner library [13] has been widely used to achieve this. Fine-tuning efficiency has been further established in the classification plant leaves disease [14], where fine-tuning architectures such as Resnet50, DenseNet121, InceptionV4, and VGG16 have been used.

Lightweight models such as MobileNetV2 and NasnetMobile have been gaining popularity recently due to the ease of their deployability. MobileNetV2[15] utilizes the concept of depth-wise separable convolution to reduce the number of training parameters without affecting the accuracy of a model. They are ideal for tasks such as recognition of palm prints [16], breast mammogram classification [17], and the identification of proper wearing of facemask [18]. Similar models like MobileNetV2 have also been built to achieve similar accuracy with fewer parameters, as in the case of PeleeNet [19]. In Deep Learning, simple transfer learning should be avoided, as it necessitates layer fine tuning [20].

## 3 Research Methodology

### 3.1 Dataset description

The dataset for this paper has been taken from the Kaggle competition [21]. There are 7049 images in this dataset and 15 facial key points representing various parts of the face such as eyebrows, eyes, nose, and lips in the training dataset. These facial key points represent the target variables. The test dataset consists of 1783 images. The dataset consists of images of 96x96 size with a one-channel dimension (grayscale images). The distribution of null values is shown below in Figure 1. 69.64% of the data points contain at least one null value in the facial key points, while 30.36% of the images consists of all key points

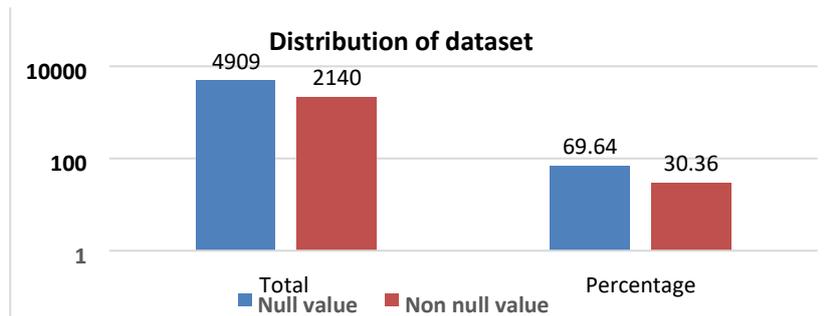

Figure 1. **Class Imbalance**

### 3.2 Image Pre-processing

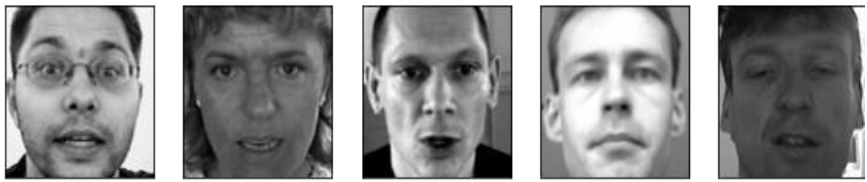

Figure 2. Visualization of Images

As mentioned below in the models' section, transfer learning architectures such as MobileNetV2 and NASNetMobile are used on this dataset along with custom-designed CNN architectures. These pre-trained networks require the input image to be in a three-channel format, and NASNetMobile requires the image size to be 224x224x3. Hence the images are converted to the appropriate format. The raw image, along with the corresponding facial key points, is shown in Figure 2.

### 3.3 Imputation techniques

**Forward fill & K-Nearest Neighbour (KNN) imputation** Forward fill is an imputation technique where the subsequent null values are filled with the previous valid observations. KNN works on imputing the missing value by predicting the nearest neighbour to a particular datapoint.

### 3.4 Data Augmentation

Figure 3 depicts a few of the augmentations used in this paper, such as random rotation, brightness, shift, and noise. These procedures were applied offline on the dataset's non-null subset.

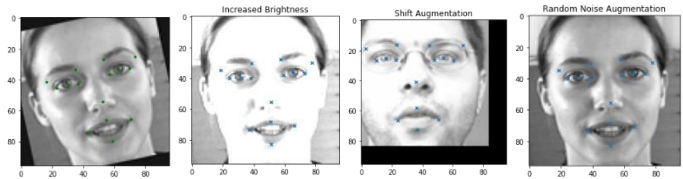

*Figure 3. Rotation, Brightness, Shift and Random noise augmentation*

### 3.5 Inference Time

The Inference times of various models have been calculated on 100 images. It can be defined as shown in Equation 1

$$Inf\ time\ on\ 100\ images = \frac{Inference\ time\ of\ the\ Total\ test\ dataset}{Number\ of\ test\ images\ in\ the\ test\ dataset}$$

Equation 1. Inference Time Calculation

### 3.6 Loss functions

The current problem is framed as a regression model where the target variable is a continuous numeric variable, the loss function used here is mean squared error. Mean squared error is defined by the following equation.

$$Mean\ Squared\ Error = \frac{1}{n}\sum_{i=1}^{n}(y_i - \hat{y}_i)^2$$

*Equation 2. Mean Squared Error - Loss function*

### 3.7 Evaluation metrics

The evaluation metric used in this regression problem is the root mean squared error (RMSE) as shown in Equation 3

$$Root\ Mean\ Squared\ Error = \sqrt{\frac{1}{n}\sum_{i=1}^{n}(y_i - \hat{y}_i)^2}$$

*Equation 3. Root Mean Squared Error*

### 3.8 Model Architecture

Two different models have been built here, Custom models and Transfer learning models, namely MobileNetV2 and NasnetMobile. Tuning is done using the Keras tuner library.

*Table 1. Parameter / Model size comparison of all architectures*

| Custom Models | Total parameters | Model size (MB) |
|---|---|---|
| Baseline CNN model | 1,890,366 | 7.6 |
| Manually Optimized CNN | 235,834 | **1.0** |
| Keras Optimized CNN- No imputation | 306,750 | 1.27 |
| Keras Optimized CNN - Forward fill | 246,478 | **1.03** |
| Keras Optimized CNN - KNN imputed | 246,062 | 1.58 |
| Keras Optimized CNN - Augmentation | 364,318 | 1.50 |
| MobilenetV2 | 2,257,984 | 9.66 |
| NasNetMobile | 4,301,426 | 18.48 |

**Custom Models.** Three different custom models have been built using baseline architecture, manual tuning, and Keras auto-tuning. The custom models are tuned sequentially to arrive at the best-performing model in terms of RMSE scores. The model's parameter count is listed in Table 1. Additionally, complete fine-tuning of transfer learning architectures was performed—the model's tuning results in a reduction in the number of parameters. Manually tuned Custom models have the least parameters with an insignificant difference in RMSE scores, as seen in Figure 6. Further, the tuned model's size is lesser than non-tuned models, with **manually tuned models having the least size** (1.0 MB). The model architecture of the manual tuned and the Keras tuned model is shown below in Figure 4.

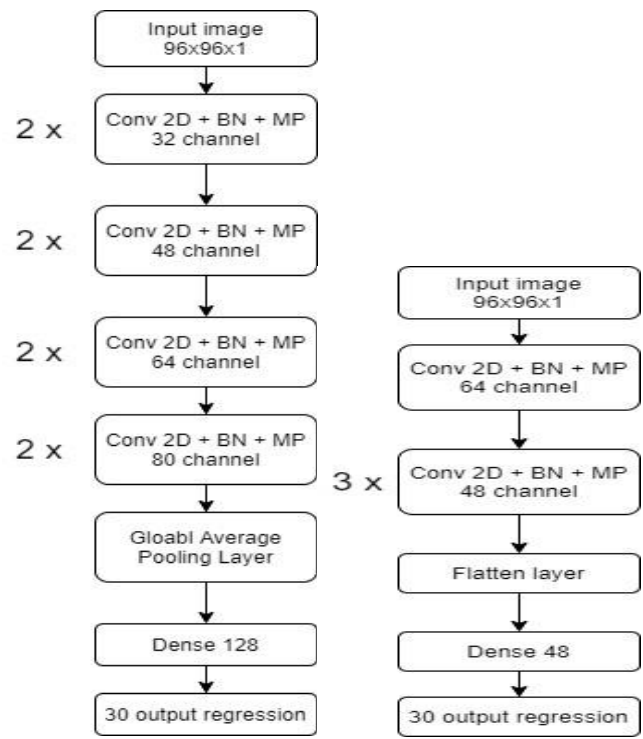

*Figure 4. Manually Tuned CNN (Left) and Keras Tuned CNN architecture (Right)*

**MobileNetV2 and NASNetMobile**. Transfer learning architectures such as MobileNetV2 and NASNetMobile have been customized to solve our regression problem. The original weights from the Imagenet classification have been used. The topmost softmax classification has been replaced with a GAP + Regression (Dense Layer) to predict the facial key points. The models are experimented with using the original baseline weights of imagenet and by completely fine-tuning all the layers of the architecture to evaluate the RMSE scores and inference time on prediction. The model architectures are shown in Figure 5

## 4 Results

The results of the experiments have been explained in the following subsections consisting of Evaluation of RMSE scores, model size, and number of parameters

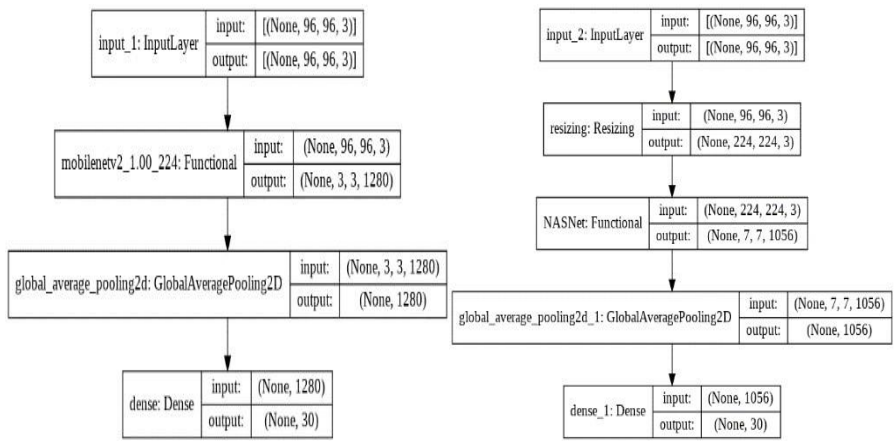

*Figure 5. MobilenetV2 (left) and NasnetMobile architecture (right)*

### 4.1 Evaluation of RMSE scores

RMSE scores on the test dataset have been calculated for both custom and transfer learning models, as shown in Figure 6 and Figure 7.

**Huge Parameters of Baseline models.** The initial baseline model was created using the conventional architecture without tuning the layers among the custom models. Figure 6 show that the Custom baseline model outperformed the manually optimized and Keras fine tuner optimized models; however, manually optimized models performed similarly to Keras fine tuner optimized models.

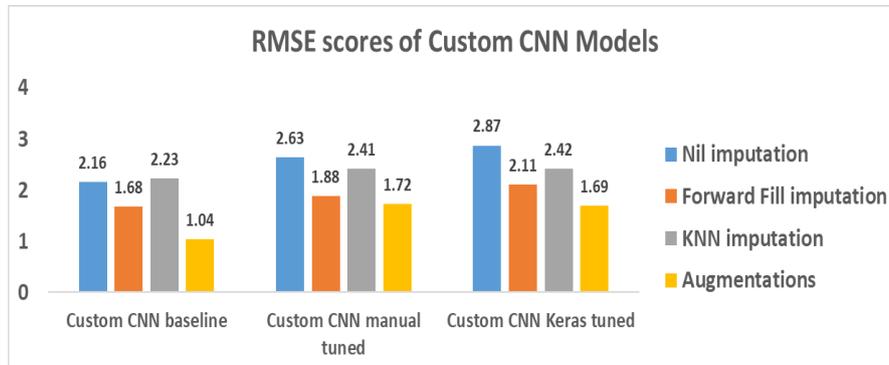

*Figure 6. RMSE scores of Custom CNN models*

It's worth noting that both fine-tuned MobileNetV2 and NASNetMobile trained on augmented data exhibit a 4-5x improvement in RMSE scores compared to their non-fine-tuned counterparts (Figure 7). Surprisingly, compared to its non-finetuned counterpart, fine-tuned MobileNetV2 demonstrated a 2x improvement in RMSE on KNN imputed data.

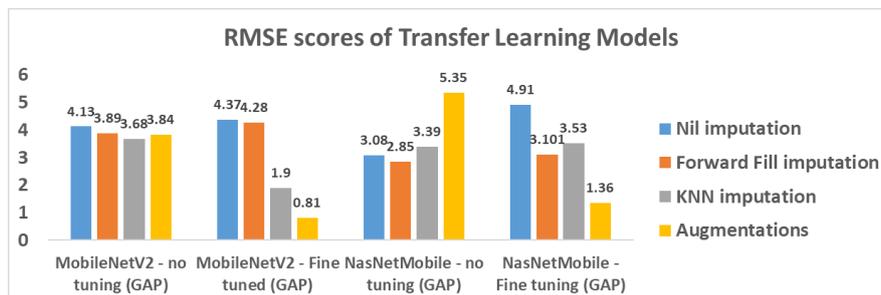

*Figure 7. RMSE scores of Transfer Learning Models*

*Table 2. Comparison of RMSE performance of transfer learning models*

| Models | No İmputation | Forward fill İmputation | KNN imputation | Aug |
|---|---|---|---|---|
| MobileNetV2 baseline model | Similar performance | Similar performance | + | + |
| MobileNetV2 fine tuned |  |  | ++ | +++ |
| NasNet baseline model | Similar performance | Similar performance | Similar performance | + |
| NasNet Model fine-tuned |  |  |  | +++ |

**Supremacy of Models Trained on Augmented Data.** As seen in Table 2 and Table 3, augmentation of custom models results in a significant increase in the performance of the models. A sharp decrease in the RMSE scores on the fine-tuned model shows that augmentation performs better than any imputation technique.

### 4.2  Evaluation of Model size and parameters

Among all the models built, manually tuned custom models have the least number of parameters (235K) and least model size against Keras auto-tuned custom models trained on different kinds of imputation techniques and augmentation (Figure 8). However, in the case of augmentation, Keras auto-tuned models slightly outperform custom models at the cost of increasing the number of parameters and model size.

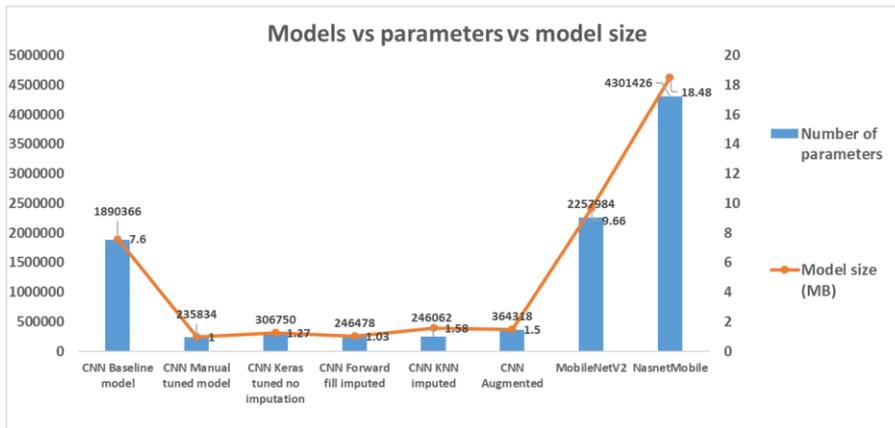

*Figure 8. Model parameters vs. Model Size – All Models*

### 4.3  Inference time analysis

*Table 3. Inference Time Analysis*

| Model | No impute (sec) | Forward Fill (sec) | KNN Impute (sec) | Aug(sec) |
|---|---|---|---|---|
| CNN Baseline model | 1.99 | 1.97 | 1.98 | 2.01 |
| **CNN Manual tuned model** | **1.4** | **1.33** | **1.34** | **1.34** |
| CNN Keras tuned | 2.72 | 1.52 | 4.19 | 3.58 |
| **MobileNetV2 - Baseline** | **0.89** | **0.86** | **1** | **0.83** |
| **MobileNetV2 - Fine tuned** | **0.84** | **0.82** | **0.82** | **0.88** |
| NasNetMobile - Baseline | 8.46 | 8.4 | 8.17 | 7.87 |
| NasNetMobile - Fine tuned | 7.95 | 7.96 | 8.01 | 7.68 |

The ultimate performance depends on the speed at which an inference can be made on the test dataset with the least computational requirements. Table 4 shows the inference time on 100 images by various models on a Colab CPU.

**Architectural efficiency in Inference Time**. The inference time of a model depends on both the number of parameters and the architecture. Among all models, MobileNetV2 has the quickest inference. The enormous training parameters (twice that of MobileNetV2) account for NASNetMobile's increased inference times. Manually tuned models come in second. In contrast to custom CNN models, MobileNetV2 has ten times the number of parameters and works two times faster. Augmentation does not affect the inference time in a regression scenario, as seen from the analysis.

### 4.4 Evaluation of Training Curves

Training curves for various models are shown below to identify the best performing model in this scenario.

**MobileNetV2 vs NasnetMobile.** Figures 9 and 10 show that NASNetMobile has better training curves than MobileNetV2 architecture for all imputation techniques and augmentation. The better training curves may be attributed to higher parameters of NASNetMobile. However, when considering inference times, RMSE scores, and parameter counts, MobileNetV2 outperforms NASNetMobile.

**Manual Tuning vs. Keras autotuning**. Manually tuned models exhibit more reliable model fitting training curves than Keras auto-tuned models, as illustrated in Figure 11.

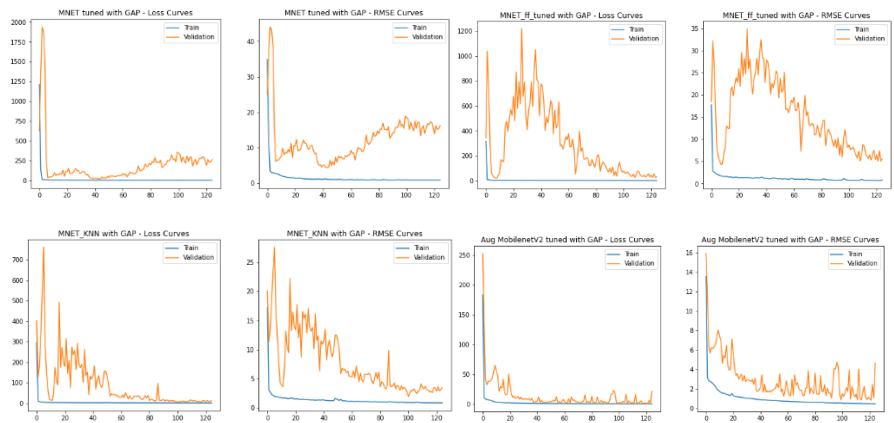

*Figure 9. MobileNetV2 - No impute, Forward Fill, KNN impute, Augmentation (Top to bottom)*

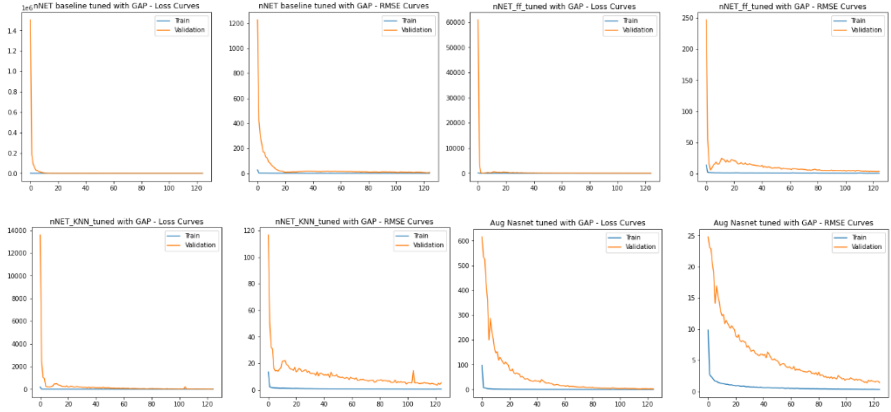

*Figure 10. NASNetMobile - No impute, Forward Fill, KNN impute, Augmentation (Top to Bottom)*

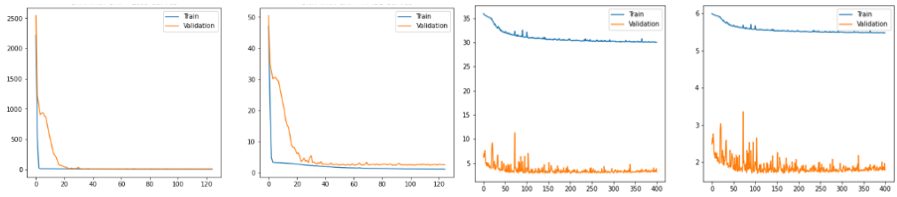

*Figure 11. Custom CNN Manual Tuned (Left) Vs Custom CNN Keras Tuned (Right)*

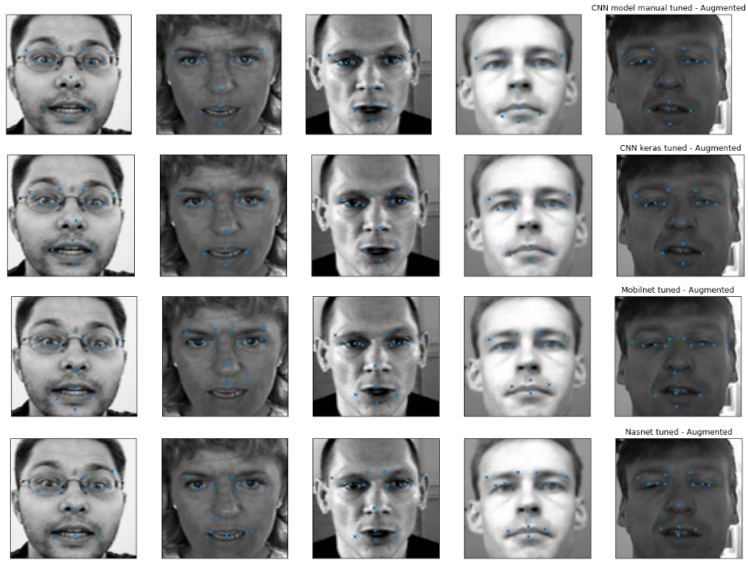

*Figure 12. Facial Key Point Predictions by CNN manual tuned/Keras auto-tuned, MobilenetV2 and NASNetMobile on Augmentation*

### 4.5 Visualization of Test Images

Figure 12 shows various augmented models' predictions of facial key points. Varying performances by different models are observed in the images below. However, the images only represent a sample of the total test dataset, and hence no meaningful conclusion can be drawn.

## 5 Conclusion

In this work, we conducted experiments on the facial key point detection dataset by building custom CNN models optimized manually and using Keras fine-tuner. Further transfer learning architectures, non-finetuned and fine-tuned MobileNetV2 and NasNetMobile were used as baselines to evaluate custom-built CNN architecture. In addition, we compared the effectiveness of imputation and augmentation. The following are the conclusions of our work which can be summarized below.

- Manually optimized custom CNN models outperform or are comparable to auto tuned Keras optimized models. On the other hand, manually tuned custom CNN models may be ideal when considering training curves, model size, and model parameters.

- MobileNetV2 outperforms all other models with the fastest inference times but slightly compromises the model size and parameters.

- In both custom CNN and transfer learning models, augmented models have lower RMSE scores, proving that augmentation is superior to imputation.

- Furthermore, there is no significant difference in performance between baseline non-tuned and baseline completely fine-tuned models, demonstrating that transfer learning models must be fine-tuned selectively in terms of the number of layers for a given dataset.

- The experiments demonstrate that architectural efficiency significantly impacts model performance and inference time, as demonstrated by the MobileNetV2 architecture, which uses depth-wise separable convolutions.

- Moreover, our models have the lowest RMSE compared to other **state-of-the-art architectures** ([4], [10], [11], [12]), and to our knowledge, this is one of the very few studies that evaluated models on size, inference time, parameters and RMSE